\documentclass{article}
\usepackage{indentfirst}
\usepackage{graphicx} 

\usepackage{wrapfig}
\usepackage[left=3cm,right=3cm,top=2cm,bottom=1.5cm]{geometry}
\usepackage{hanging}
\usepackage{ragged2e,array}
\newcolumntype{P}[1]{>{\centering\arraybackslash}p{#1}}

\usepackage{amsmath}
\usepackage{authblk}
\usepackage{url}
\usepackage[hyphenbreaks]{breakurl}
\usepackage{underscore}
\usepackage{xurl}

\title{Pre-Ictal Seizure Prediction Using Personalized Deep Learning}

\author[1]{Shriya Jaddu}
\author[2]{Sidh Jaddu}
\author[3]{Camilo Gutierrez}
\author[1,4,5]{Quincy K. Tran}

\affil[1]{Research Associate Program in Emergency Medicine and Critical Care, Department of Emergency Medicine, University of Maryland School of Medicine, Baltimore, Maryland.}
\affil[2]{Thomas Jefferson High School for Science and Technology, Alexandria, Virginia.}
\affil[3]{Department of Neurology, University of Maryland School of Medicine, Baltimore, Maryland.}
\affil[4]{Department of Emergency Medicine, University of Maryland School of Medicine, Baltimore, Maryland.}
\affil[5]{The R Adam Cowley Shock Trauma Center, University of Maryland School of Medicine, Baltimore, Maryland.}

\date{}

\begin{document}
\maketitle
\begin{center} \section*{Abstract} \end{center}

\subsection*{Introduction}
Approximately 23 million or 30\% of epilepsy patients worldwide suffer from drug-resistant epilepsy (DRE). The unpredictability of seizure occurrences, which causes safety issues (swimming, driving, etc.) as well as social concerns (gatherings, etc.), restrict the lifestyles of DRE patients. Surgical solutions such as deep brain stimulation that minimize seizure occurrences are very expensive, unreliable, and invasive. Current electroencephalogram (EEG)-based solutions that predict seizures are impractical for 24-hour use. Although few researchers have attempted to use physiological data to predict seizures during the pre-ictal period in order to remove the barriers faced by DRE patients, they have not had acceptable model performances for practical patient use. The goal of this research was to employ improved technologies and methods to epilepsy patients' physiological data and predict seizures up to two hours before onset, enabling non-invasive, affordable seizure prediction for DRE patients.

\subsection*{Methods}

This research used a 1D Convolutional Neural Network-Based Bidirectional Long Short-Term Memory (CNN-BiLSTM) network that was trained on a diverse set of epilepsy patients’ physiological data to predict seizures. Transfer learning was further utilized to personalize and optimize predictions for specific patients. Clinical data was retrospectively obtained for nine epilepsy patients collected via wearable devices over  periods of about three to five days from a prospectively maintained database. The physiological data collected included 54 seizure occurrences, and it included the variables heart rate, blood volume pulse, accelerometry, body temperature, and electrodermal activity.

\subsection*{Results}
A general deep-learning model trained on epileptic patient physiological data with randomly sampled test data achieved an accuracy of 91.94\%. However, such a generalized deep learning model had varied performances on data from unseen patients. When the general deep learning model was personalized (further trained) with patient-specific data, the personalized model achieved significantly improved performance for the patient. With personalization, the models achieved prediction accuracies as high as 97\%.

\subsection*{Conclusion}
Although deep learning models trained with physiological data offer the ability to predict upcoming seizures, personalization of these models showed significantly improved prediction accuracies. This preliminary research shows that patient-specific personalization may be a viable approach to achieve affordable, non-invasive seizure prediction that can improve the quality of life for DRE patients.

\vspace{\baselineskip} 
\vspace{\baselineskip} 

\section*{Introduction}

Epilepsy is a neurological disorder characterized by recurrent seizures, episodes of abnormal neuron activity. Epilepsy can be caused by a variety of underlying conditions, including genetic disorders, brain damage, etc; however, 50\% of epilepsy cases do not correlate with any known underlying condition at all (World Health Organization, 2024). Thus, because epilepsy is a diverse neurological disorder, symptoms experienced by patients before seizures (during the pre-ictal period) and during seizures (during the ictal period) can vary greatly (CDC, 2020). 

\vspace{\baselineskip}

Epilepsy affects more than seventy million people worldwide (Löscher et al., 2020). Although there are more than forty medications available to prevent seizure events, one-third of epilepsy patients worldwide (about 23 million patients) suffer from drug-resistant epilepsy (DRE), causing them to have uncontrolled seizures. During seizure events, patients may zone out, fall unconscious, lose control of parts of their body, and may have bodily injuries. After their seizures, patients may also experience a period of disorientation that can last anywhere from a few minutes to a few hours. Because of these symptoms as well as the unpredictability of seizures, DRE patients often lead restricted lifestyles due to safety and social issues (Löscher et al., 2020). Furthermore, repeated and prolonged seizures can sometimes have long-term adverse effects on brain functions in certain patients. Therefore, predicting and alerting upcoming seizure activity in epileptic patients is crucial in improving these patients’ quality of life. 

\vspace{\baselineskip} 

Currently, there are a few solutions to DRE seizures. For some patients, surgery can be performed to remove the affected part of the brain to stop the seizures. Alternatively, vagus nerve stimulators and deep brain stimulators are potential solutions to prevent or reduce seizures (Freund et al., 2021). However, these solutions are invasive, expensive (costing anywhere from \$30,000 to \$200,000), and inaccessible to patients in different parts of the world, especially in developing countries where access to advanced treatment and diagnosis is limited (Singh et al., 2022; New Choice Health, 2023).

\vspace{\baselineskip} 

In recent years, Artificial Intelligence (AI) has been gaining popularity and is being applied to solve various problems in the medical field. Researchers have begun to explore the potential of using AI to support individuals with epilepsy. Some products in the market such as Empatica Embrace2 and Epilepsy Foundation’s SmartWatch utilize AI to aid epilepsy patients; however, these devices are designed only to detect ongoing seizures during the ictal stage, rather than to predict upcoming seizures during the pre-ictal stage (Empatica, 2024; Epilepsy Foundation, 2024). 

\vspace{\baselineskip} 

Several researchers have been exploring applications of AI to electroencephalography (EEG) data to predict seizures through two common methods of EEG collection: subcutaneous EEG and ear-EEG devices. Subcutaneous EEG requires invasive implants into the brain, making it impractical for many epilepsy patients. Similarly, ear-EEG devices, while non-invasive because they can be worn on the ears, are suboptimal, and, based on previous studies, patients using ear-EEG devices reported discomfort and discontinued use, deeming it unsuitable for 24-hour use (Zibrandtsen et al., 2017). 

\vspace{\baselineskip} 

Researchers have begun to propose that an effective approach to DRE is through seizure prediction during the pre-ictal period using patient physiological data that is collected non-invasively. Research has shown that patients can experience changes in physiological features such as blood volume pulse, electrodermal activity, heart rate, etc. during the pre-ictal period. Although researchers have attempted to train AI on physiological data to predict seizures during the pre-ictal period, acceptable model performances for practical patient use have not been achieved.

\vspace{\baselineskip} 

This research sought to more accurately predict seizures up to a few hours before they occur by utilizing physiological data and personalized deep learning algorithms. Accurate prediction of these seizures allows patients to take necessary precautions such as taking an additional dose of medication or modifying their activities.

\section*{Methods}

\subsection*{Data Acquisition and Collection}

Physiological seizure data was acquired from the Epilepsy Foundation of America’s My Seizure Gauge Public Dataset (Kuhlmann et al., 2018; Nasseri et al., 2020). The dataset included data from nine epilepsy patients whose physiological data was recorded for three to five-day periods. Data was collected from patients both in in-patient and in-home settings using Empatica E4 wristwatches, which collected data from five physiological sensors (Nasseri et al., 2020). The dataset used for modeling includes data collected from blood volume pulse (BVP), electrodermal activity (EDA), heart rate (HR), accelerometry (ACC), and temperature (TEMP) sensors. Blood volume pulse indicates blood flow and is measured as the heart beats. Electrodermal activity broadly refers to the electrical activity of the skin but also includes characteristics such as perspiration. Accelerometry is a measure of motion and acceleration: in the dataset, the accelerometry data included data from three channels (X, Y, Z) as well the magnitude of these channel values. Only the magnitude was utilized in the final dataset.

\vspace{\baselineskip} 

In order to build deep learning models for binary classification of pre-ictal and interictal physiological data, segments of pre-ictal and interictal data for all 9 patients were selected for training and testing from the original My Seizure Gauge Public Dataset. The selected dataset included data from 54 seizures of various types and symptoms including generalized tonic-clonic, clonic, focal (motor, complex, and discognitive), myoclonic, hypermotor, and subclinical. Relatively equal amounts of interictal and pre-ictal data were selected to prevent major class imbalances in subsequent modeling. The number of seizures and minutes of pre-ictal data from each patient are shown in Table 1 below.

\begin{table}[!h] 
\renewcommand{\arraystretch}{2}
\centering
\begin{tabular}{| >{\Centering}m{1in} | >{\Centering}m{1in} | >{\Centering}m{.75in} | >{\Centering}m{1.2in} | >{\Centering}m{1.2in} |}
\hline
\textbf{Patient ID} & \textbf{Seizure Types} & \textbf{Number of Seizures} & \textbf{Total Pre-Ictal Data (min)} & \textbf{Total Interictal Data (min)} \\
\hline
01575 & Focal & 3 & 130 & 130 \\
\hline
00172 & Clonic, Hypermotor & 4 & 480 & 480 \\
\hline
00501 & Focal, Clonic & 3 & 75 & 75 \\
\hline
01097 & Focal, GTC & 4 & 128 & 135 \\
\hline
01110\_ICU & Focal (Complex), GTC & 5 & 108 & 107 \\
\hline
01808 & Myoclonic, GTC, Focal (Discognitive) & 4 & 58 & 55 \\
\hline
01838 & GTC, Focal & 9 & 234 & 232 \\
\hline
01842 & Focal (Discognitive), GTC & 8 & 116 & 114 \\
\hline
01844 & Focal (Motor) & 14 & 76 & 76 \\
\hline
\textbf{Total} &  & \textbf{54} & \textbf{1405 minutes (23.4 hours)} & \textbf{1404 minutes (23.4 hours)} \\
\hline

\end{tabular}

\end{table}

\begin{center} \text{Table 1. Number of seizures and amount of data (minutes) collected per patient}  \end{center}

 \subsection*{Data Preprocessing}

The data was scaled using robust data scaling. The scaled values were calculated using Equation 1 as shown below. In Equation 1, X represents the original value, X\textsubscript{median} represents the median of all of the data points, X\textsubscript{new }represents the scaled value, and IQR is the interquartile range. 

 \begin{figure}[!h]
     \centering
     \includegraphics[width=0.25\linewidth]{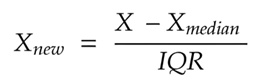}
 \end{figure}

\begin{center} 
\text{Equation 1} \end{center}
\vspace{\baselineskip} 

The median value of all of the data points was calculated and subtracted from the original data value, and the resulting value was divided by the interquartile range, which is defined as the range between the 25\% and 75\% percentiles of the data. Robust scaling not only reduced the impact of outliers but also, considering that all of the data points were originally extremely large numbers, helped with the efficiency of the modeling process.

\vspace{\baselineskip} 

A sample duration of 30 seconds was chosen in this research to ensure that training samples were long enough for models to accurately analyze and understand patterns in the data. Thus, as 30-second segments of pre-ictal and interictal (normal) data were used for training and testing, the dataset was broken down into arrays of 30 seconds of data for the time-series classification model. The format of a single training sample is shown in Figure 1 below.

\begin{figure}[!h]
    \centering
    \includegraphics[width=1\linewidth]{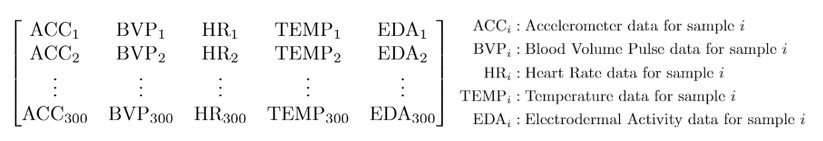}
    \caption{Format of a Single Training Sample}
\end{figure}

Finally, three exclusive datasets were created randomly for modeling: a training dataset (60\% of the original data), a validation dataset (20\% of the original data), and a testing dataset (20\% of the original data). The datasets were cross-checked thoroughly to ensure they did not contain overlapping samples.

\subsection*{Deep Learning Algorithm}
Several multivariate (comprising all five sensors) model architectures including Bidirectional Long Short-Term Memory (BiLSTM), Convolutional Neural Network Long Short-Term Memory (CNN-LSTM), and Convolutional Neural Network-based Bidirectional Long Short-Term Memory (CNN-BiLSTM) networks were experimented with. 

\vspace{\baselineskip} 

LSTM models, typically used for time-series forecasting or classification, are a subset of Recurrent Neural Networks that focus on developing an understanding of time-series data by storing particular information from input data. BiLSTM models traverse training data in both the forward and reverse directions, improving the model's performance (Siami-Namini et al., 2020). Convolutional Neural Network-based LSTMs can capture spatial features and understand patterns in time-series data (Mohammed \& Corzo, 2024). Finally, a CNN-BiLSTM architecture combines the convolution layers of the CNN-LSTM architecture and the bidirectionality of the BiLSTM architecture, allowing it to capture spatial features in the data and traverse input data in the forward and backward directions. The performance metrics of these models are shown in Table 2.

\vspace{\baselineskip}
\vspace{\baselineskip}
\vspace{\baselineskip}
\vspace{\baselineskip}
\vspace{\baselineskip}

\begin{table}[!h] 
\centering

\renewcommand{\arraystretch}{2}
\begin{tabular}{| >{\Centering}m{1.4in} | >{\Centering}m{.7in} | >{\Centering}m{.7in} | >{\Centering}m{.7in} | >{\Centering}m{.7in} | >{\Centering}m{.8in} |}
\hline
\textbf{Model Architecture} & \textbf{Accuracy} & \textbf{F1 Score} & \textbf{Precision} & \textbf{Recall} & \textbf{AUC-ROC} \\
\hline
BiLSTM & 83.03\% & 0.7927 & 0.7947 & 0.7921 & 0.7921 \\
\hline
CNN-LSTM & 85.27\% & 0.8509 & 0.8603 & 0.8495 & 0.8495 \\
\hline
CNN-BiLSTM & 91.94\% & 0.9079 & 0.9377 & 0.8800 & 0.9159 \\
\hline

\end{tabular}

\end{table}

\begin{center}  \text{Table 2. Model Performance Metrics for Different Deep Learning Architectures}  \end{center}

\vspace{\baselineskip} 

Figure 2 shows a model architecture of a CNN-BiLSTM model.

\begin{figure}[!h] 
    \centering 
    \includegraphics[width=1.0\linewidth]{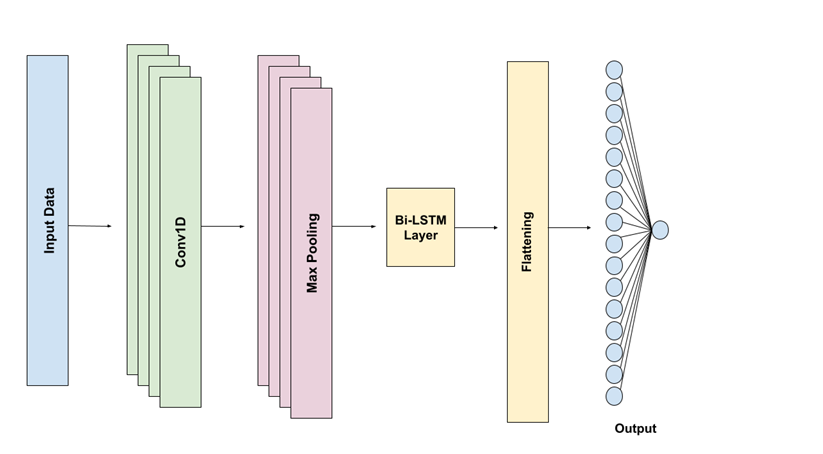}
    \caption{Model Architecture of a CNN-BiLSTM Model 
}
    \label{fig:enter-label}
\end{figure}

Single variate models were also trained with data from each sensor to understand their performances. However, the multivariate model outperformed all of the single variate models. 

\vspace{\baselineskip} 

The selected CNN-BiLSTM multivariate model architecture included three 1D Convolutional layers, each of which utilized a Rectified Linear Unit (ReLU) activation function. Following each of the three Convolutional layers, 1D Max Pooling was employed to extract the most significant features. Next, the model incorporated a Bidirectional LSTM layer and a flattening layer, which flattened the output of the BiLSTM layer into a 1D vector. Finally, two dense layers were implemented, both of which utilized sigmoid activation, considering that sigmoid activation is suitable for binary classification tasks. Hyperparameters were selected after extensive tuning. The model was trained on the training dataset for 10 epochs with a batch size of 32. The model was validated using the already-created validation set and evaluated using the binary cross-entropy loss function. The model utilized the Adam optimizer with a learning rate of 0.001. The Keras callback ReduceLRonPlateau was implemented to reduce the learning rate if no improvement was seen in the model after 3 epochs. 

\vspace{\baselineskip} 

Additionally, class weights were implemented to correct any class imbalances between the pre-ictal and interictal data using Equation 2 (as shown below):

\vspace{\baselineskip} 
\vspace{\baselineskip} 
\vspace{\baselineskip} 

 \begin{figure}[!h] 
     \centering
     \includegraphics[width=0.45\linewidth]{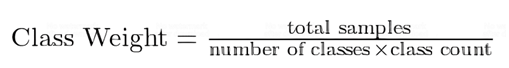}
 \end{figure}
 
\begin{center} 
\text{Equation 2} \end{center}
\vspace{\baselineskip}

In Equation 2 above, the total samples was defined as the number of samples in the training dataset, the number of classes was two (considering that the model was carrying out binary classification), and the class count was the number of samples per class in the training dataset.

\vspace{\baselineskip} 

A “general” model was trained and tested with the diverse data of the patients. To confirm the efficacy of the general model for unseen patients, the general model was tested on data from unseen patients. The general model did not perform very well when tested for unseen patients. This is because patients may have different pre-ictal patterns, making it challenging for a single general model to effectively predict seizures for all seizure patients. 

\vspace{\baselineskip} 

To combat this, personalization was implemented using a transfer learning-like technique. Personalization, in this research, refers to the optimization of the general model’s performance for a certain patient. In the implementation of personalization, the general model was initially trained on all of the patient data except for the data from the patient on whom personalization was being implemented. Then, this general model was tested on the patient to be personalized, and the performance metrics before personalization were captured. As the next step, the general model was then used as a base model, and the specific patient data (which was excluded from training the general model) was used to train the model further. The patient-specific data samples were given additional weight compared to the base data, to ensure that the model was maintaining past learnings from the general pool of data, while also optimizing to fit the patient-specific patterns. This personalized model was then tested on the data of that patient and the performance metrics after personalization were captured. This process was repeated for each of the patients and the corresponding before and after personalization performance metrics were calculated.

\subsection*{Performance Metrics}
Model performance was evaluated using common deep learning metrics including accuracy, F1 score, precision, recall, and area under the receiver operating characteristic (AUC-ROC). Equations 3-6 for the metrics are as follows:

\begin{figure}[!h]
    \centering
    \includegraphics[width=0.7\linewidth]{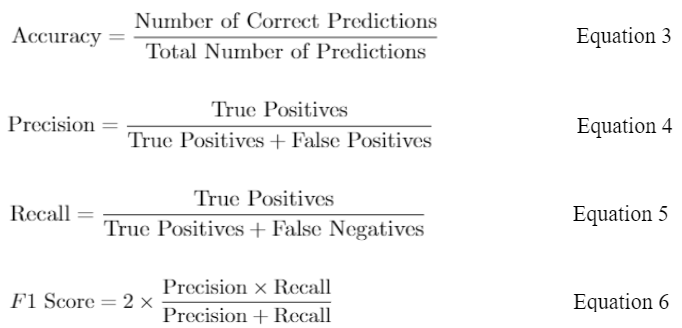}
\end{figure}

Accuracy provides a general indication of how many correct predictions the model made without considering class imbalances. The precision represents the positive predictive value of the model performance. The recall represents the true positive rate. The F1 score represents the harmonic mean of precision and recall and is a good measure of performance when there is a class imbalance. The AUC-ROC score reports how well a model performs for both the positive and negative classes by calculating the area under the receiver operating characteristic curve, which plots the model performance over various classification thresholds.

\section*{Results and Discussion}
Several model architectures were considered in developing the general model, including CNN-LSTM and BiLSTM. However, these architectures underperformed in comparison to the CNN-BiLSTM, hence the CNN-BiLSTM was ultimately chosen (Table 2).

\vspace{\baselineskip} 

The general multivariate model, which was trained on data from all nine patients, achieved an accuracy of 91.94\%. The performance metrics of the general multivariate model are shown in Table 3 below.

\vspace{\baselineskip}

\begin{table}[!h]
\centering

\renewcommand{\arraystretch}{2}
\begin{tabular}{| >{\Centering}m{1in} | >{\Centering}m{1in} | >{\Centering}m{1in} | >{\Centering}m{1in} | >{\Centering}m{1in} |}
\hline
\textbf{Accuracy} & \textbf{Precision} & \textbf{Recall} & \textbf{F1 Score} & \textbf{AUC-ROC} \\
\hline
91.94\% & 0.9377 & 0.8800 & 0.9079 & 0.9159 \\
\hline

\end{tabular}

\end{table}

\begin{center} \text{Table 3. Performance Metrics of the General Physiological Model} \end{center} 

\vspace{\baselineskip} 

The model generalized and performed well on a balanced set of pre-ictal and interictal segments, considering that both the precision and recall were high. The confusion matrix and the ROC curve for the general model are shown in Figure 3 below.  

 \begin{figure}[!h]
     \centering
     \includegraphics[width=1\linewidth]{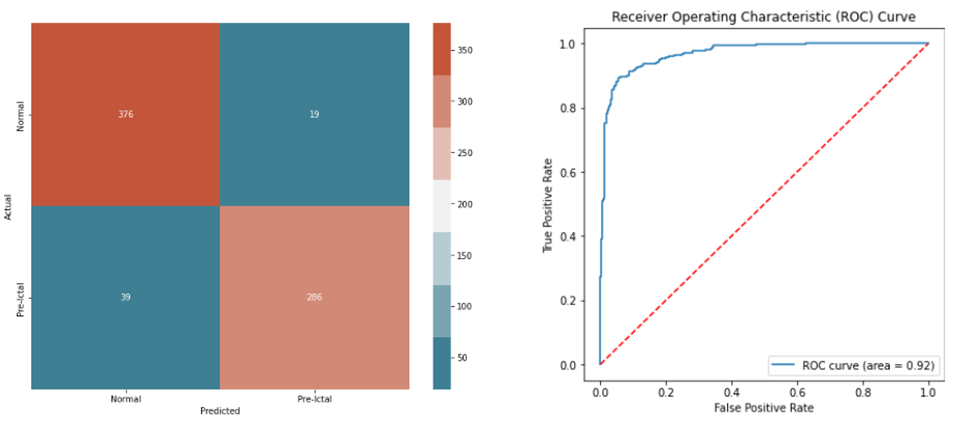}
 \end{figure}

\begin{center} \text{Figure 3. Confusion Matrix and ROC Curve for the General Model} \end{center} 

 However, when the general model was tested on the data of each patient separately, it did not perform well for some patients. Table 4 shows model performance on specific patients before and after personalization was implemented. For example, as shown in Table 4 below, the generalized model predicted one patient’s test data with an accuracy of 65.22\% without personalization. However, with personalization, the model predicted with an accuracy of 94.78\%. Similarly, the generalized model predicted another patient’s test data with an accuracy of 50\%. However, with personalization, the model predicted with an accuracy of 90\%. Thus, the patient-specific personalization was effective at addressing the issue of different patients having different pre-ictal patterns. These increases in performances were seen after training patient-specific models with pre-ictal data from as few as two seizure events. Appendices 1 and 2 include the precision, recall, and F1 scores before and after personalized training was implemented.

\vspace{\baselineskip}
\vspace{\baselineskip} 
\vspace{\baselineskip} 
\vspace{\baselineskip} 
\vspace{\baselineskip} 
\vspace{\baselineskip} 
\vspace{\baselineskip}  
\vspace{\baselineskip} 
\vspace{\baselineskip} 
\vspace{\baselineskip} 

\begin{table}[!h]
\centering
\renewcommand{\arraystretch}{2}
\begin{tabular}{| >{\Centering}m{1.2in} | >{\Centering}m{2.2in} | >{\Centering}m{2.2in} | }
\hline
\textbf{Patient ID} & \textbf{Patient Accuracy Before Personalization (Tested on General Model)} & \textbf{Patient Accuracy After Personalization (Tested on Personalized Model)} \\
\hline
01838 & 73.91\% & 91.30\% \\
\hline
01575 & 65.70\% & 97.09\% \\
\hline
00501 & 50.00\% & 90.00\% \\
\hline
01097 & 55.26\% & 92.11\% \\
\hline
00172 & 65.22\% & 94.78\% \\
\hline
01110-ICU & 35.15\% & 90.10\% \\
\hline
01808 & 60.00\% & 90.00\% \\
\hline
01842 & 31.31\% & 92.99\% \\
\hline
01844 & 56.74\% & 80.34\% \\
\hline

\end{tabular}

\end{table}

\begin{center}
\text{Table 4. Model Performance on Specific Patients Before and After Personalization} 
\end{center}

\vspace{\baselineskip} 

The models were trained and tested to predict the seizures as early as two hours before they occurred. The personalized model performance metrics exceeded previous studies that have attempted to use physiological data for seizure prediction.

\vspace{\baselineskip} 

Overall, the research confirmed that using the pre-training deep learning models combined with patient personalization i.e., personalized model training can significantly improve model performance.

\subsection*{Limitations and Future Work}
The dataset used for training and testing included physiological data spanning 3 to 5 days for 9 epileptic patients. The dataset was relatively small, and more data needs to be collected to evaluate the personalization mechanism further. Additionally, while this dataset encompassed various seizure types, it lacked representation from atonic and absence seizures. So, testing and validation of predictions for these seizure types could not be performed. Additionally, the model data was not categorized by demographics, hence modeling and testing did not consider demographic factors. However, because of the personalization component of the system, it is expected to predict well for all demographic groups even if a demographic group was not represented in the general model’s training data. Additionally, in order for the system to be practical for real-world use, the precision must be very high (close to 1) for the false alarm rate to be very low. This may be achieved with additional training data or other approaches (e.g., meta-learning). While the research demonstrated noteworthy performance of models after personalized training, it necessitates further validation through the use of additional patient physiological data as well as clinical trials before it is brought into the real world.

\section*{Conclusion}
This research presents preliminary findings that show that personalized deep learning models trained with physiological seizure data can provide significantly improved prediction accuracies for seizure prediction during the pre-ictal period. Accurate prediction of upcoming seizures will enable DRE patients to take precautionary measures and improve their quality of life.

\pagebreak
\begin{center}
\section*{References}
\end{center}

\begin{hangparas}{.25in}{1}

\textit{Antiseizure medications (Formerly known as anticonvulsants)}. (2023, February 3). Cleveland Clinic. Retrieved December 24, 2023, from \url{https://my.clevelandclinic.org/health/treatments/24781-antiseizure-medications-anticonvulsants}

Embrace2. (n.d.). Retrieved January 5, 2024, from \url{https://www.empatica.com/
embrace2/}

Faru, S. H., Waititu, A., \& Nderu, L. (2023). A hybrid neural network model based on transfer learning for forecasting forex market. Journal of Data Analysis and Information Processing, 11(02), 103-120. \url{https://doi.org/10.4236/jdaip.2023.112007}

Freund, B., Feyissa, A., Grewal, S., Middlebrooks, E., \& Sirven, J. (n.d.). Neurostimulation in drug-resistant epilepsy. \textit{Practical Neurology}. \url{https://practicalneurology.com/articles/2021-oct/neurostimulation-in-drug-resistant-epilepsy}

Hamad, R., Yang, L., Woo, W., \& Wei, B. (2020, July). Architecture of the hybrid 1D CNN + LSTM model for human activity recognition. [Chart]. ResearchGate. 
\url{https://www.researchgate.net/publication/343341551_Joint_Learning_of_Temporal_Models_to_Handle_Imbalanced_Data_for_Human_Activity_Recognition}

Kashouty. (n.d.). \textit{Understanding the different types of epileptic seizures and how they manifest}. Premier Neurology and Wellness Center. Retrieved December 24, 2023, from 
\url{https://premierneurologycenter.com/blog/understanding-the-different-types-of-epileptic-seizures-and-how-they-manifest/}

Kuhlmann, L., Karoly, P., Freestone, D. R., Brinkmann, B. H., Temko, A., Barachant, A., Li, F., Titericz, G., Jr, Lang, B. W., Lavery, D., Roman, K., Broadhead, D., Dobson, S., Jones, G., Tang, Q., Ivanenko, I., Panichev, O., Proix, T., Náhlík, M., Grunberg, D. B., … Cook, M. J. (2018). Epilepsyecosystem.org: crowd-sourcing reproducible seizure prediction with long-term human intracranial EEG. \textit{Brain: a journal of neurology, 141}(9), 2619–2630. \url{https://doi.org/10.1093/brain/awy210}

Livingstone S., Russo F. (2018) The Ryerson Audio-Visual Database of Emotional Speech and Song (RAVDESS): A dynamic, multimodal set of facial and vocal expressions in North American English. PLoS ONE 13(5): e0196391. \url{https:// doi.org/10.1371/journal.pone.0196391.}

Löscher, W., Potschka, H., Sisodiya, S. M., \& Vezzani, A. (2020). Drug resistance in epilepsy: Clinical impact, potential mechanisms, and new innovative treatment options. \textit{Pharmacological Reviews, 72}(3), 606-638. \url{https://doi.org/10.1124/pr.120.019539}

Nasseri, M., Nurse, E., Glasstetter, M., Böttcher, S., Gregg, N. M., Laks Nandakumar, A., Joseph, B., Pal Attia, T., Viana, P. F., Bruno, E., Biondi, A., Cook, M., Worrell, G. A., Schulze-Bonhage, A., Dümpelmann, M., Freestone, D. R., Richardson, M. P., \& Brinkmann, B. H. (2020). Signal quality and patient experience with wearable devices for epilepsy management. \textit{Epilepsia, 61 Suppl 1,} S25–S35. \url{https://doi.org/10.1111/epi.16527}

Mohammed, A., \& Corzo, G. (2024). Spatiotemporal convolutional long short-term memory for regional streamflow predictions. Journal of Environmental Management, 350, 119585. \url{https://doi.org/10.1016/j.jenvman.2023.119585}

National Center for Chronic Disease Prevention and Health Promotion, \& Division of Population Health. (2020, September 30). \textit{Types of seizures}. Centers for Disease Control and Prevention. Retrieved December 24, 2023, from \url{https://www.cdc.gov/epilepsy/about/types-of-seizures.htm}

Pandya, V., \& Carlson, C. (n.d.). Epilepsy essentials: Neuromodulation for drug-resistant epilepsy. \textit{Practical Neurology}. \url{https://practicalneurology.com/articles/2021-june/epilepsy-essentials-neuromodulation-for-drug-resistant-epilepsy}

Rudzicz, F., Namasivayam, A.K., Wolff, T. (2012) The TORGO database of acoustic and articulatory speech from speakers with dysarthria. Language Resources and Evaluation, 46(4), pages 523--541.

Salanova, V., Witt, T., Worth, R., Henry, T. R., Gross, R. E., Nazzaro, J. M., Labar, D., Sperling, M. R., Sharan, A., Sandok, E., Handforth, A., Stern, J. M., Chung, S., Henderson, J. M., French, J., Baltuch, G., Rosenfeld, W. E., Garcia, P., Barbaro, N. M., . . . Epstein, C. (2015). Long-term efficacy and safety of thalamic stimulation for drug-resistant partial epilepsy. Neurology, 84(10), 1017-1025. \url{https://doi.org/10.1212/wnl.0000000000001334}

Scaramelli, A., Braga, P., Avellanal, A., Bogacz, A., Camejo, C., Rega, I., Messano, T., \& Arciere, B. (2009). Prodromal symptoms in epileptic patients: Clinical characterization of the pre-ictal phase. Seizure, 18(4), 246-250. \url{https://doi.org/10.1016/j.seizure.2008.10.007}

\textit{Seizure}. (2022, April 13). Cleveland Clinic. Retrieved December 24, 2023, from \url{https://my.clevelandclinic.org/health/diseases/22789-seizure}

Selerity. (2020, December 18). Techniques of feature scaling with SAS custom macro. Selerity. Retrieved December 24, 2023, from \url{https://seleritysas.com/2020/12/18/ techniques-of-feature-scaling-with-sas-custom-macro/}

Siami-Namini, S., Tavakoli, N., \& Namin, A. S. (2020). The performance of LSTM and BiLSTM in forecasting time series. IEEE International Conference on Big Data, 3285-3292. \url{https://doi.org/10.1109/BigData47090.2019.9005997}

Singh, R., Zamanian, C., Bhandarkar, A., Shoushtari, A., Bydon, M., Gottfried, O., Southwell, D., Gendreau, J., Brown, N., \& Shahrestani, S. (2022). \textit{Predictors of increased cost in epilepsy surgery} [Paper presentation]. Congress of Neurological Surgery Annual Meeting, San Francisco. \url{https://www.researchgate.net/publication/362509419\_Predictors\_of\_Increased\_Cost\_in\_Epilepsy\_Surgery}

SmartWatch by Smart Monitor. (2022, March 11). Retrieved January 5, 2024, from
\url{https://www.epilepsy.com/deviceapedia/smartwatch-smart-monitor-0}

Tang, J., El Atrache, R., Yu, S., Asif, U., Jackson, M., Roy, S., Mirmomeni, M., Cantley, S., Sheehan, T., Schubach, S., Ufongene, C., Vieluf, S., Meisel, C., Harrer, S., \& Loddenkemper, T. (2021). Seizure detection using wearable sensors and machine learning: Setting a benchmark. \textit{Epilepsia}, \textit{62}(8), 1807-1819. \url{https://doi.org/10.1111/epi.16967}

Tao, J. (2022, December 5). \textit{New treatment options for people with drug-resistant epilepsy}. UChicago Medicine. Retrieved December 23, 2023, from 
\url{https://www.uchicagomedicine.org/forefront/neurosciences-articles/new-treatment-options-for-people-with-drug-resistant-epilepsy}

Téllez-Zenteno, J. F., Dhar, R., \& Wiebe, S. (2005). Long-term seizure outcomes following epilepsy surgery: A systematic review and meta-analysis. Brain, 128(5), 1188-1198. \url{https://doi.org/10.1093/brain/awh449}

Tasdelen, A., Sen, B. A hybrid CNN-LSTM model for pre-miRNA classification. Sci Rep 11, 14125 (2021). \url{https://doi.org/10.1038/s41598-021-93656-0}

\textit{Understanding vagus nerve stimulator (VNS) placement}. (n.d.). Saint Luke's. Retrieved December 24, 2023, from \url{https://www.saintlukeskc.org/health-library/understanding-vagus-nerve-stimulator-vns-placement}

\textit{Vagus nerve stimulation cost and vagus nerve stimulation procedures information}. (n.d.). New Choice Health. Retrieved December 24, 2023, from \url{https://www.newchoicehealth.com/vagus-nerve-stimulation-cost}

World Health Organization. (2024, February 7). Epilepsy. Retrieved June 14, 2024, from \url{https://www.who.int/news-room/fact-sheets/detail/epilepsy}

Zibrandtsen, I., Kidmose, P., Christensen, C., \& Kjaer, T. (2017). Ear-EEG detects ictal and interictal abnormalities in focal and generalized epilepsy – A comparison with scalp EEG monitoring. \textit{Clinical Neurophysiology}, \textit{128}(12), 2454-2461. \url{https://doi.org/10.1016/j.clinph.2017.09.115}

\end{hangparas}

\pagebreak
\textbf{Appendix 1. }General model performance on data of each of the nine patients before personalized training is implemented.

\begin{table}[!h]
\centering

\renewcommand{\arraystretch}{2}
\begin{tabular}{| >{\Centering}m{1in} | >{\Centering}m{1.1in} | >{\Centering}m{1.1in} | >{\Centering}m{1.1in} | >{\Centering}m{1in} |}
\hline
\textbf{Patient ID} & \textbf{Accuracy} & \textbf{Precision} & \textbf{Recall} & \textbf{F1 Score} \\
\hline
01838 & 73.91\% & 1.000 & 0.7391 & 0.8500 \\
\hline
01575 & 65.70\% & 0.4928 & 0.8870 & 0.6335 \\
\hline
00501 & 50.00\% & 0.5000 & 0.4000 & 0.4444 \\
\hline
01097 & 55.26\% & 0.5556 & 0.6034 & 0.5785 \\
\hline
00172 & 65.22\% & 1.000 & 0.6522 & 0.7895 \\
\hline
01110-ICU & 35.15\% & 0.3604 & 0.4000 & 0.3791 \\
\hline
01808 & 60.00\% & 0.5606 & 0.7708 & 0.6491 \\
\hline
01842 & 31.31\% & 0.3504 & 0.4528 & 0.3951 \\
\hline
01844 & 56.74\% & 0.4346 & 0.7381 & 0.5471 \\
\hline

\end{tabular}

\end{table}

\vspace{\baselineskip}
\vspace{\baselineskip}
\vspace{\baselineskip}
\vspace{\baselineskip}

\textbf{Appendix 2. }Model performance on each of the nine patients after personalized training is implemented.

\begin{table}[!h]
\centering

\renewcommand{\arraystretch}{2}
\begin{tabular}{| >{\Centering}m{1in} | >{\Centering}m{1.1in} | >{\Centering}m{1.1in} | >{\Centering}m{1.1in} | >{\Centering}m{1in} |}
\hline
\textbf{Patient ID} & \textbf{Accuracy} & \textbf{Precision} & \textbf{Recall} & \textbf{F1 Score} \\
\hline
01838 & 91.30\% & 1.000 & 0.9130 & 0.9545 \\
\hline
01575 & 97.09\% & 0.9817 & 0.9304 & 0.9554 \\
\hline
00501 & 90.00\% & 1.000 & 0.8000 & 0.9000 \\
\hline
01097 & 92.11\% & 0.9455 & 0.8966 & 0.9204 \\
\hline
00172 & 94.78\% & 1.000 & 0.9478 & 0.9732 \\
\hline
01110-ICU & 90.10\% & 0.8774 & 0.9300 & 0.9029 \\
\hline
01808 & 90.00\% & 0.9318 & 0.8542 & 0.8913 \\
\hline
01842 & 92.99\% & 0.9789 & 0.8774 & 0.9254 \\
\hline
01844 & 80.34\% & 0.8043 & 0.5873 & 0.6789 \\
\hline

\end{tabular}

\end{table}

\end{document}